\documentclass{article}

\usepackage{PRIMEarxiv}

\usepackage[utf8]{inputenc} 
\usepackage[T1]{fontenc}    
\usepackage{hyperref}       
\usepackage{url}            
\usepackage{booktabs}       
\usepackage{amsfonts}       
\usepackage{nicefrac}       
\usepackage{microtype}      
\usepackage{lipsum}
\usepackage{fancyhdr}       
\usepackage{graphicx}       
\graphicspath{{media/}}     

\title{Have Multimodal Large Language Models (MLLMs) Really Learned to Tell the Time on Analog Clocks?
}

\author{
 Tairan Fu \\
  College of Mechanical and Electrical Engineering\\
  Nanjing University of Aeronautics and Astronautics\\
  Nanjing, China \\
  \texttt{tkm199888@gmail.com} \\
   \And
 Miguel González, Javier Conde, Pedro Reviriego \\
  ETSI de Telecomunicación\\
  Universidad Politécnica de Madrid\\
  Madrid, Spain \\
  \texttt{\{miguel.gonsaiz,javier.conde.diaz,pedro.reviriego\}@upm.es} \\
  \And
 Elena Merino-Gómez \\
  Escuela de Ingenierías Industriales\\
  Universidad de Valladolid\\
  Madrid, Spain  \\
  \texttt{elena.merino.gomez@uva.es} \\
}

\begin{document}
\maketitle

\begin{abstract}
Multimodal Large Language Models which can answer complex questions on an image struggle to tell the time on analog clocks. This is probably due to the lack of images with clocks at different times in their training set. In this work we explore this issue with one of the latest MLLMs: GPT-4.1 to understand why MLLMs fail to tell the time and whether fine-tuning can solve the problem. The results show how models are making progress in reading the time on analog clocks. But have they really learned to do it, or have they only learned patterns in their training datasets? In this work we put the models to the test with different clocks to illustrate the limitations of MLLMs to abstract and generalize.
\end{abstract}


\section{Introduction}
With the tremendous success of large language models (LLMs)\cite{1}, research and applications have expanded into higher-dimensional modalities such as images and audio, giving rise to MLLMs, which have already demonstrated success across various domains\cite{2}\cite{3}. However, it is worth noting that, taking image processing tasks as an example, although many MLLMs can handle sophisticated image understanding tasks like visual question answering\cite{4}, most of them surprisingly fail at a seemingly simple task: telling the time on an analog clock\cite{5}.

\begin{figure}
\centerline{\includegraphics[width=18.5pc]{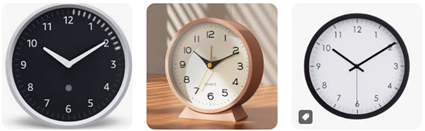}}
\caption{Examples of clock images found in an internet search}\vspace*{-5pt}
\label{fig:1}
\end{figure}

A similar phenomenon was previously observed in LLMs when processing text data. For instance, early LLMs often failed to count letters correctly or perform basic arithmetic. However, these issues can be significantly mitigated by increasing the scale of the training datasets\cite{6}\cite{7}. By analogy, the failure of MLLMs to read clocks can likely be attributed to the lack of diverse \{image, text\} pairs of clocks at different times in their training datasets. Indeed, most of the datasets are extracted from the Internet and do not cover all possible times and are highly biased towards some specific times like for example 10:10 as shown in Figure \ref{fig:1}.

\begin{figure*}[!t]
\centering
\begin{minipage}[b]{0.45\linewidth}
  \centering
  \includegraphics[width=\linewidth]{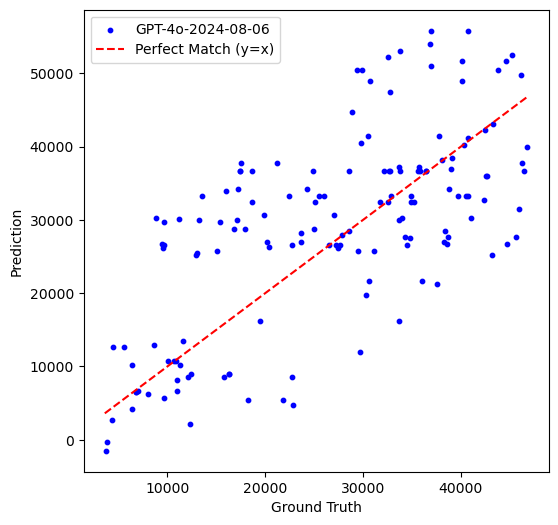}
  \par\vspace{1pt}
       {\fontsize{8}{8}\selectfont (a) Before fine-tuning}
\end{minipage}
\hfill
\begin{minipage}[b]{0.45\linewidth}
  \centering
  \includegraphics[width=\linewidth]{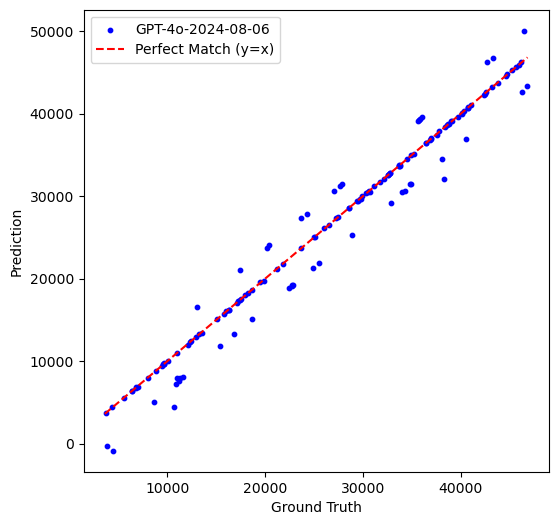}
    \par\vspace{1pt}
     {\fontsize{8}{8}\selectfont (b) After fine-tuning}
\end{minipage}
\caption{Effect of LLM fine-tuning on model prediction performance}
\label{fig:2}
\end{figure*}

To investigate this hypothesis, we have constructed a dataset\footnote{\url{https://huggingface.co/datasets/migonsa/analog\_watches\_finetune}} of clock images and corresponding times that cover all possible time options and then used some of those pairs to fine-tune a MLLM, one of the initial versions of GPT-4o\footnote{The version used was gpt-4o-2024-08-06}. Prior to fine-tuning, the model consistently failed to tell the time on these clocks. After fine-tuning, however, the model was able to accurately tell the time on previously unseen clocks within the synthetic dataset\footnote{The results for all the experiments mentioned in the paper and a more detailed analysis of the errors in the time estimation are available at \url{https://github.com/aMa2210/MLLMs\_read\_analog\_clocks}}, as shown in Figure \ref{fig:2}. This figure shows the predicted versus real time in seconds for 150 random times with the lowest value, 3600 corresponding to 1:00:00 and the highest value, 46799 to 12:59:59. The red dashed line corresponds to the ideal results where predictions are equal to the real value. It can be observed that before the fine tuning the values are scattered and deviate significantly from the correct results. After the fine tuning, some errors still persist but the predicted times are much better aligned to the correct results. This is a strong indication that the inability of GPT-4o to read clocks is due to the lack of relevant examples in its training dataset. However, if those examples were present in the dataset, what would MLLMs really learn? 

\section{Have MLLMs really learned to tell the time?}
One of the latest models of OpenAI available at the time of writing this brief: GPT-4.1\footnote{The exact model was gpt-4.1-2025-04-14} was able to read time on our synthetic clock dataset, probably because more examples of clocks have been included in its training dataset. A question of interest is whether these additional examples in the training dataset are sufficient for MLLMs to learn to tell the time, or whether they are just learning the patterns in those examples\cite{8}. To explore this, we have also created clock images with unusual shapes and changed the width and shape of the hands. Interestingly, the use of deformed clocks is one of the recurring themes in Salvador Dalí’s paintings such as the persistence of memory\footnote{\href{https://www.moma.org/collection/works/79018}{Salvador Dalí. The Persistence of Memory. 1931 | MoMA}} and the Spanish artist surname was an inspiration for the name of one of the most well-known text to image models: DALL-E\footnote{\href{https://time.com/collections/the-ai-dictionary-from-allbusiness-com/7273943/definition-of-dall-e/}{The Definition of DALL-E}}.  A person would be able to tell the time in these clocks with no difficulty, but would MLLMs behave like us? 

\begin{figure*}
\centerline{\includegraphics[width=26pc]{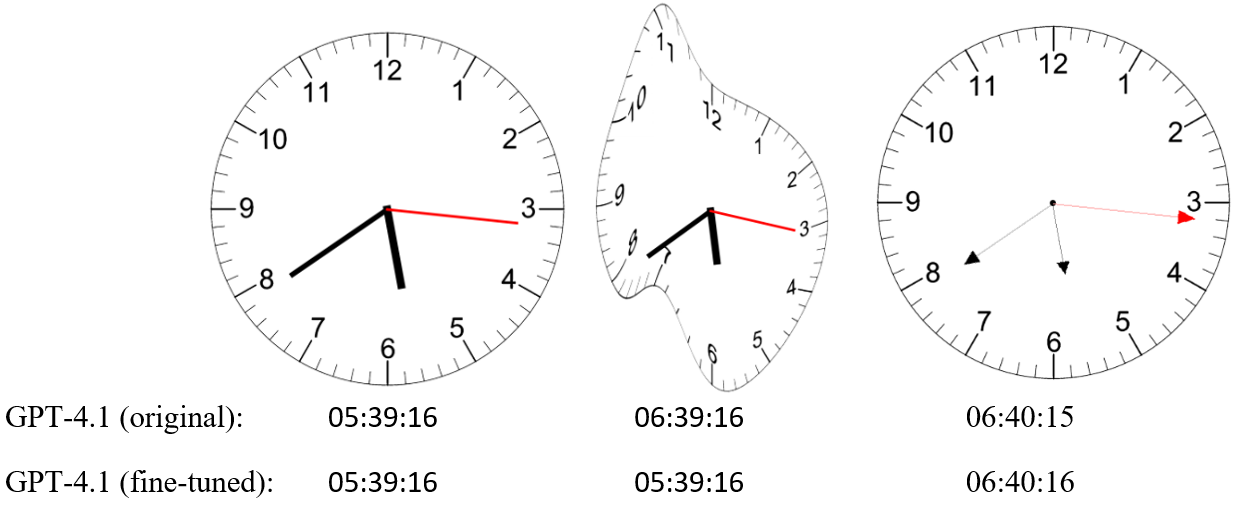}}
\caption{Normal (left), distorted shape (middle) and modified hands (right) clock images and the times told by the original and fine-tuned versions of GPT-4.1 in each case}\vspace*{-5pt}
\label{fig:3}
\end{figure*}

Examples of the initial and modified clocks are shown in Figure \ref{fig:3} with the corresponding times told by GPT-4.1, it can be seen how the model is able to read the time in the initial clock; however, it fails when the clock is deformed or when the hands are changed to be thinner and to have an arrowhead. The Mean Absolute Error (MAE) in the time estimate over 150 random times was 232.48s for the initial clocks, 1380.69s when the shape is deformed and 3726.93s when hands are changed. These results suggest that the MLLM has not learned to tell the time but rather memorized patterns in the training set. To further explore this limitation, we fine-tuned the model with another set of 300 random samples from each of the datasets and measured the error again on the 150 examples. The MAEs are now 33.82s, 784.41s and 2893.57s which show that the model achieves a higher relative improvement on clocks it already performed well on, yet struggles to effectively transfer previously learned patterns to other types of clocks.

\section{What factors contribute to MLLM's failure in reading analog clocks?}
Although GPT-4.1 performs exceptionally well with standard clock images, it is surprising that modifying the clock hands by making them thinner and adding arrowheads leads to a significant drop in its accuracy. Intuitively, one might expect that the more visually complex change—a distorted dial—would have a greater impact on performance, yet this modification seems to have a relatively smaller effect.

This raises a question: how do MLLMs interpret clocks, and why do they fail? One possibility is that thinner hands impair the model's ability to perceive direction, weakening its understanding of spatial orientation. Alternatively, there could be other factors that cause confusion when the model attempts to combine the hour, minute, and second hands into an accurate time reading.
\begin{figure*}
\centerline{\includegraphics[width=26pc]{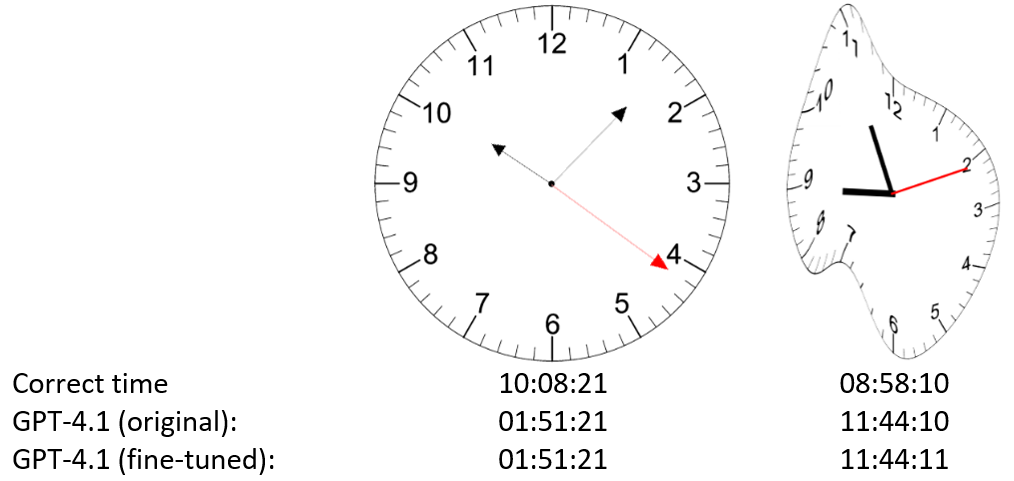}}
\caption{Examples of hand function confusion: Modified hands (left) and distorted shape (right) clocks misread by GPT-4.1 before and after fine-tuning}\vspace*{-5pt}
\label{fig:4}
\end{figure*}

Understanding the underlying cause of this issue is crucial for the future development of MLLMs. If the problem is related to the model's sensitivity to spatial directions, then further fine-tuning could quickly address this limitation. On the other hand, if the problem arises from an inability to effectively coordinate multiple pieces of information, it would indicate a deeper issue with the model's ability to integrate and process information.

To further investigate this issue, we examined the results. As expected, we observed that the majority of error cases are caused by misinterpretations of the clock hands' directions. As shown in Figure \ref{fig:2}, this issue accounts for all the errors in the normal clock dataset and the majority of errors in the distorted shape clock dataset. In contrast, with the modified hands clock dataset, we identified a different issue: the model confuses the functions of the clock hands. This issue also appears in the distorted shape clock dataset, although it occurs less frequently. Figure \ref{fig:4}. shows two examples of the hour and minute hands being confused by the model and resulting in incorrect time readings. 

This indicates two inherent issues with MLLMs on this task. First, there is a lack of sensitivity to directional perception. Second, the model seems to suffer from overfitting, probably due to insufficient diversity in the training data, leading it to overly rely on the thickness of the clock hands to determine their types.

\section{Understanding the impact of the two factors}
Although it was previously noted that MLLMs perform poorly on the modified hands clock dataset both before and after fine-tuning, the same model performs well on the normal clock dataset, and fine-tuning further improves its performance. This discrepancy prompts further investigation into whether the performance drop is solely due to the confusion about the functions of the clock hands. If the model correctly distinguishes the hands, would fine-tuning be as effective as it is on the normal clock dataset?

\begin{table}
  \caption{Analysis of samples with and without clock hand function confusion}
  \centering
  \begin{tabular}{llccc}
    \toprule
    Hand type & Error type & Number & MAE (Before fine-tuning) & MAE (After fine-tuning) \\
    \midrule
    Modified & Confused & 45 & 10088.9s & 7937.1s \\
             & Not confused & 99 & 882.2s & 595.8s \\
    Normal   & Confused & 0 & N.A. & N.A. \\
             & Not confused & 150 & 232.5s & 33.8s \\
    \bottomrule
  \end{tabular}
  \label{table1}
\end{table}

\begin{table}
  \caption{Analysis of samples with and without clock hand function confusion}
  \centering
  \begin{tabular}{lllcc}
    \toprule
    Hand type & Error type & Hand Role & MAE (Before fine-tuning) & MAE (After fine-tuning) \\
    \midrule
    Modified & Confused & Hour & 84.0° & 66.7° \\
             &          & Minute & 80.9° & 58.7° \\
             &          & Second & 24.9° & 8.4° \\
             & Not confused & Hour & 7.3° & 4.5° \\
             &              & Minute & 5.8° & 6.7° \\
             &              & Second & 4.8° & 1.0° \\
    Normal   & Not confused & Hour & 2.4° & 0° \\
             &              & Minute & 1.6° & 1.0° \\
             &              & Second & 1.4° & 1.9° \\
    \bottomrule
  \end{tabular}
  \label{table2}
\end{table}

To explore this, we conducted a more detailed analysis of the MLLM's results on the modified hands clock dataset. Based on whether the model correctly identifies the functions of the clock hands, the samples were divided into two categories. Notably, six samples were excluded from the analysis because the reason of the error could not be clearly determined. We then calculated the MAEs of GPT-4.1 for both categories before and after fine-tuning, alongside its performance on normal clocks, as shown in Table \ref{table1}. To further analyze the composition of the errors, we calculated the individual errors of the hour, minute, and second hands for each data group. To ensure readability and consistency of the data, we used their corresponding angles on the clock dial as the unit. The results are shown in Table \ref{table2}.

These results indicate that confusion regarding the function of the hands is the primary source of error in the model’s performance. This is understandable, as mistaking the hour and minute hands can lead to errors of several hours, whereas directional misjudgment typically results in discrepancies of at most one or two hours. Besides, as shown in Table \ref{table2}, the model exhibits the largest angular error on the hour hand and the smallest on the second hand before fine-tuning, which may be attributed to the different lengths of the hands, leading to varying difficulties in direction estimation.

To better understand the model’s unexpected underperformance, we excluded cases involving hand function confusion and focused on the remaining examples. Ideally, the model should have achieved performance comparable to that on the normal clock dataset—or at least shown a clear improvement after fine-tuning. However, the results indicate that the model still performed noticeably worse on these samples.

One possible explanation is that the altered shape of the clock hands may interfere with the model’s perception of direction. To test this hypothesis, we created two sets of 60 synthetic images, each containing only an hour hand pointing to every minute mark on the dial—one set with normal hands and another with modified hands. The MLLM was then asked to answer which tick mark [0–59] the hand was pointing to. The model achieved similar performance on both sets, with MAEs of 6.5° and 8.1° after converting the results into angular measurements. While the modified hands resulted in slightly higher error, the difference is too small to fully account for the substantial performance drop observed when reading complete clocks. This suggests that when the model is confused by one aspect of the input, its performance in other areas—where it would otherwise excel—can also deteriorate.

\begin{figure}
\centerline{\includegraphics[width=18.5pc]{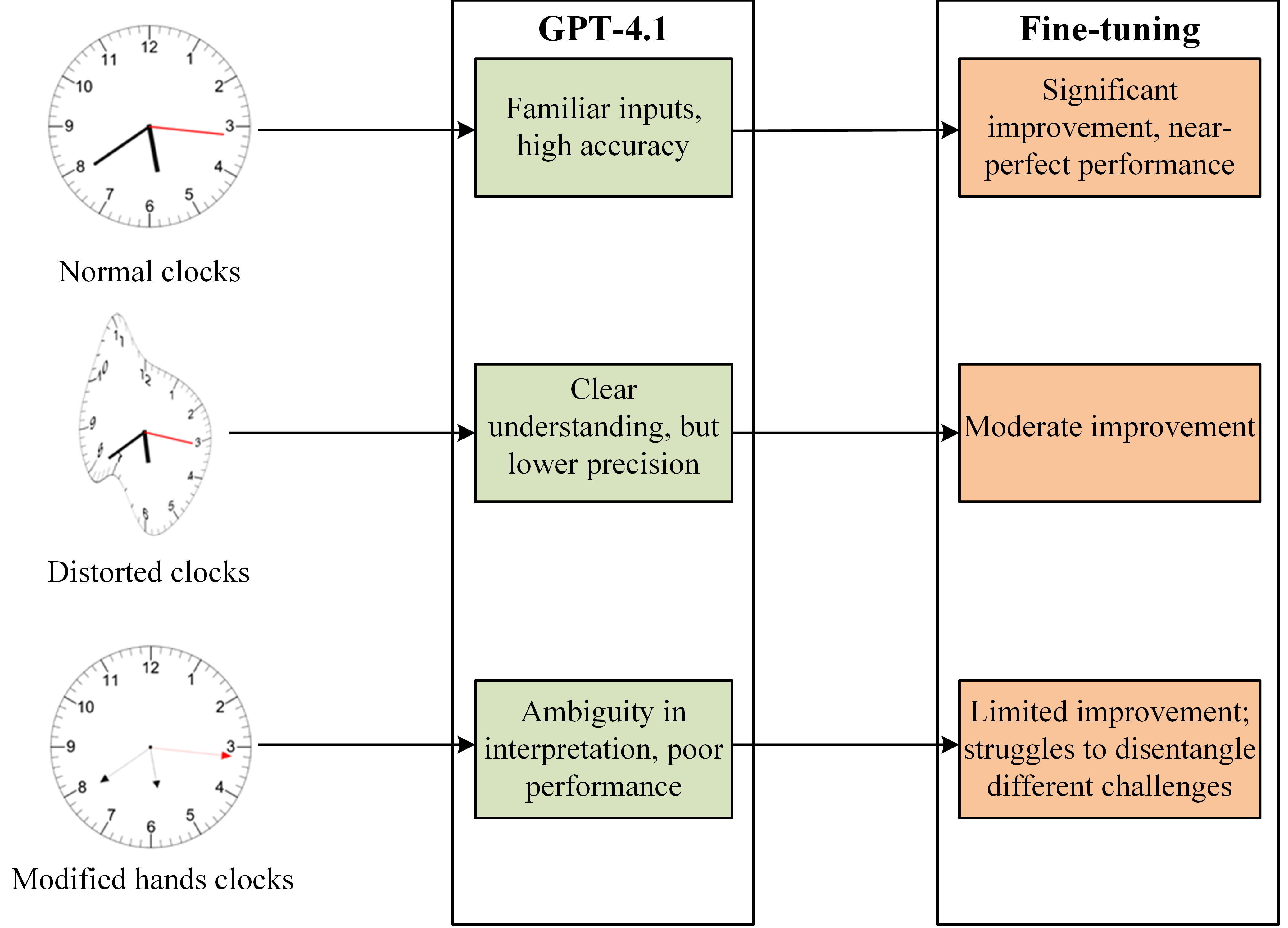}}
\caption{Performance of GPT-4.1 across clock variants and fine-tuning impact}\vspace*{-5pt}
\label{fig:5}
\end{figure}

Additionally, the effect of fine-tuning appears similarly limited under such interference. The model’s error on this subset decreased by only 32.5\%, which is less than the reduction observed on both the dataset where the model originally performed worse (deformed clocks, with a 43.2\% decrease) and one where it performed better (normal clocks, with an 85.5\% decrease). This points to a second limitation of fine-tuning: it becomes less effective when the model is exposed to multiple sources of interference. An overview of the MLLM's performance across different clock types is summarized in Figure \ref{fig:5}.

This may reveal a potential bottleneck for MLLMs when facing increasingly complex multimodal information in the future. While MLLMs exhibit strong performance on familiar inputs, their overall capabilities can degrade significantly once local changes interfere with the understanding of key elements. More importantly, such performance degradation is often difficult to detect. For example, even when the model correctly identifies the type of clock hands, increased uncertainty may still weaken its ability to perceive their direction. In more complex real-world scenarios—such as medical image analysis or autonomous driving perception—these subtle yet critical failures could lead to more severe consequences. Even more concerning is the fact that fine-tuning does not appear to be an efficient method for such latent issues. This calls for a reevaluation of current training paradigms for MLLMs and highlights the need for more structured approaches to improve their generalization and robustness in complex environments.

\section{Discussion and limitations}
The ability of the model to tell the time in the Dali’s inspired distorted clocks is suggestive and connects with the wordplay given that GPT-4.1 is developed by OpenAI which also designed DALL-E. Beyond this anecdote, the example of the clocks illustrates the limitations of current MLLMs in generalizing simple tasks thar are trivial for humans\cite{8}. This can be addressed by enlarging the training datasets with more and more examples, but this does not seem to be scalable solution, having to fix each new case with a new addition of elements to the training dataset. Therefore, the challenge is to develop models which can learn at higher levels of abstraction, capable of adapting to changes in patterns as we humans do.

Although this paper analyzes the challenges faced by current MLLMs in reading analog clocks, there are still many limitations. By using the task of reading analog clocks as a case study, this paper pointed out that fine-tuning may not be the most efficient solution. However, for such simple tasks, constructing a dataset and continuing fine-tuning is undoubtedly still the most economical approach. Also it is foreseeable that companies will improve MLLMs so they will soon have the capability to accurately and reliably tell the correct time based on analog clock images. Nevertheless, for many other types of problems, constructing large datasets may not be as simple as generating synthetic clocks. In this context, analyzing the shortcomings of MLLMs in fine-tuning, as pointed out in this paper, are an interesting area for further research.

\section*{Acknowledgments}
The authors thank Javier Coronado for his valuable comments and insights. This work was supported by the Spanish Agencia Estatal de Investigación under Grants FUN4DATE (PID2022-136684OB-C22) and SMARTY (PCI2024-153434) and SMARTY funded by the European Commission through the Chips Act Joint Undertaking project SMARTY (Grant 101140087).

\bibliographystyle{unsrt}  
\bibliography{references}

\end{document}